\documentclass[letterpaper, 10 pt, conference]{ieeeconf}

\usepackage{cite} 
\usepackage{multirow}
\usepackage{hyperref}
\usepackage[utf8]{inputenc} 
\usepackage[misc]{ifsym}
\usepackage{graphicx}
\usepackage{tikz}
\usepackage{algorithm}

\usepackage{algorithmic}
\usepackage{pifont}
\newcommand{\cmark}{\ding{51}}%
\newcommand{\xmark}{\ding{53}}%

\usepackage[misc]{ifsym}
\usepackage{tikz}
\usepackage{pdfpages}
\usepackage{graphicx}
\usepackage[utf8]{inputenc}
\usetikzlibrary{arrows,shapes,positioning,shadows,trees,calc,decorations.markings}

\usepackage{amssymb}

\IEEEoverridecommandlockouts                              

\title{\LARGE \bf
AP-MTL: Attention Pruned Multi-task Learning Model for Real-time Instrument Detection and Segmentation in Robot-assisted Surgery
}

\author{Mobarakol Islam, Vibashan VS and Hongliang Ren
\thanks{Authors are with the Medical Mechatronics Lab, Dept. of Biomedical Engineering, National University of Singapore, Singapore- http://bioeng.nus.edu.sg/mm, Corresponding author, Hongliang Ren, ren@nus.edu.sg.}
}

\usepackage{amsmath}
\begin{document}

\maketitle
\thispagestyle{empty}
\pagestyle{empty}

\begin{abstract}
 
Surgical scene understanding and multi-tasking learning are crucial for image-guided robotic surgery. Training a real-time robotic system for the detection and segmentation of high-resolution images provides a challenging problem with the limited computational resource. The perception drawn can be applied in effective real-time feedback, surgical skill assessment, and human-robot collaborative surgeries to enhance surgical outcomes. For this purpose, we develop a novel end-to-end trainable real-time Multi-Task Learning (MTL) model with weight-shared encoder and task-aware detection and segmentation decoders. Optimization of multiple tasks at the same convergence point is vital and presents a complex problem. Thus, we propose an asynchronous task-aware optimization (ATO) technique to calculate task-oriented gradients and train the decoders independently. Moreover, MTL models are always computationally expensive, which hinder real-time applications. To address this challenge, we introduce a global attention dynamic pruning (GADP) by removing less significant and sparse parameters. We further design a skip squeeze and excitation (SE) module, which suppresses weak features, excites significant features and performs dynamic spatial and channel-wise feature re-calibration. Validating on the robotic instrument segmentation dataset of MICCAI endoscopic vision challenge, our model significantly outperforms state-of-the-art segmentation and detection models, including best-performed models in the challenge.
\end{abstract}
\section{Introduction} 
 Image-guided robotic surgery is the latest form of development in minimally invasive surgical technology, which increases precision, reliability, and repeatability. A Vision-enabled surgical system like Da Vinci \cite{ngu2017vinci} is eminently assisting the surgeon in performing complex surgical manipulations with minimal incisions and recovery time than conventional surgeries. Despite development in surgical technology, sometimes surgeons may lose surgical workflow due to weak tactile feedback or system impairment. Moreover, a compound, surgical scenario with smoke, body fluid, blood, adverse lighting condition, and partial occlusion creates additional challenges in image cognition \cite{pakhomov2017deep}. Therefore, concurrent detection and segmentation of instruments could enhance surgical outcomes and assist novice surgeons with real-time objective feedback, do skill assessment, and analyze tool movements in the surgical workflow.

In recent years, deep neural network is showing success in the tasks of segmentation \cite{ronneberger2015u}, detection\cite{redmon2018yolov3} and visual perception \cite{seo2019visualizing} from computer vision to medical imaging. However, most of these models are designed and optimized for the specific task. These restrain the impact of the deep learning model full exploitation, whereas the human brain is capable of doing multiple tasks like classification, tracking, and reasoning in parallel. Therefore, developing a multi-task learning (MTL) model in surgical scene understanding with instrument detection and segmentation can play an important rule to advance the robotic intervention. 

\begin{figure*}[!h]
\centering
\includegraphics[width=.8\textwidth]{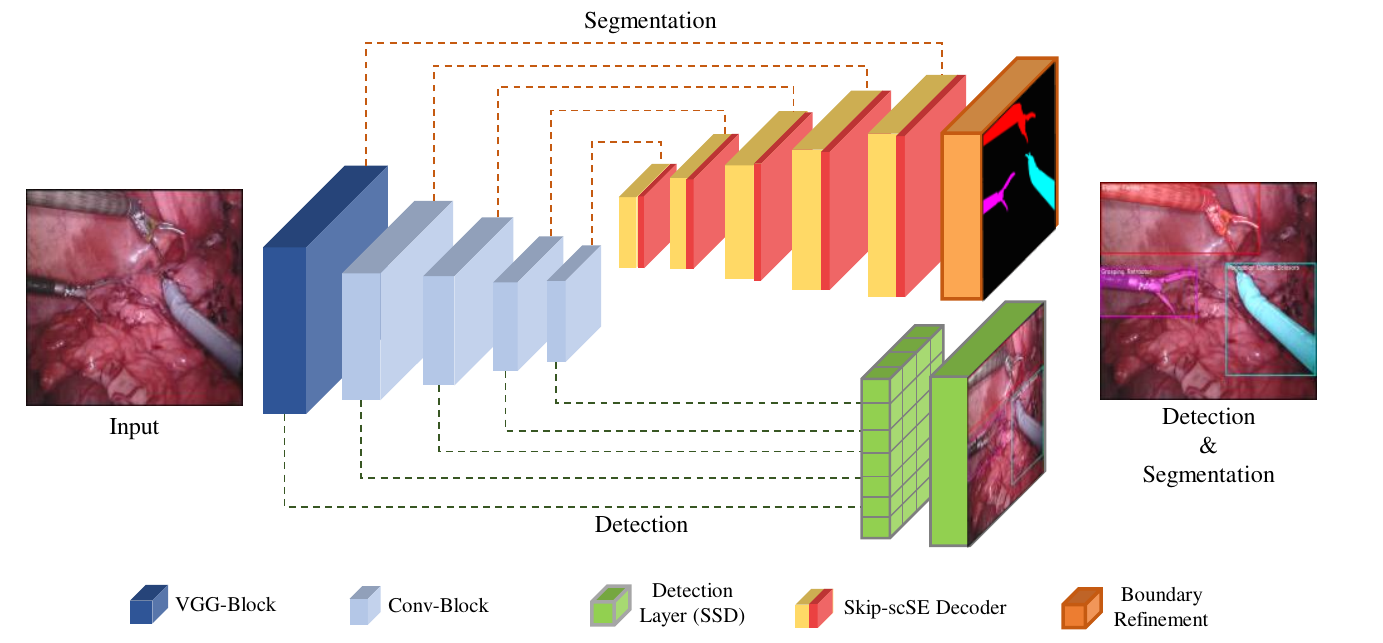}
\caption{Network architecture of the proposed AP-MTL model for real-time detection and segmentation. The architecture consists of VGG16 encoder, segmentation decoder like \cite{islam2019learning}, and detection branch like SSD \cite{liu2016ssd}.}
\label{fig:net_arch}
\end{figure*} 
 
\subsection{Related Work}

\subsubsection{Surgical Instrument Segmentation}
Recently, CNN models are deploying to achieve state-of-the-art performance in surgical tool tracking. Instrument segmentation of binary, parts and type-wise is done by LinkNet \cite{chaurasia2017linknet} with Jaccard index based loss function \cite{shvets2018automatic}. Subsequently, ToolNet \cite{garcia2017toolnet} with holistically-nested architecture, and dilated residual network with multi-scaled inputs \cite{pakhomov2017deep} are utilized to segment the surgical tool. Nonetheless, these works lack focusing on the real-time application in terms of both speed and accuracy. A real-time instrument segmentation approach of using auxiliary supervised adversarial learning is proposed in \cite{islam2019real}. The work demonstrates high-speed inference with satisfying accuracy in binary instrument segmentation. However, it performs poorly for the type-wise segmentation. In our previous work \cite{islam2019learning}, a novel design of segmentation decoder with spatial and channel squeeze \& excitation (scSE)\cite{roy2018concurrent} is improved the instrument type segmentation by exciting the significant features.

\subsubsection{Surgical Instrument Detection}
There are few works with CNN on surgical instrument tracking by detection. R-CNN with region proposal networks (RPN) \cite{ren2015faster} is exploited to detection the instrument in robot-assisted surgery \cite{sarikaya2017detection}. Subsequently, a residual CNN \cite{chen2017surgical}, and RANSAC based CNN \cite{zhao2017tracking} are used to detect bounding box with the tooltip. A 3D fully connected network (FCN) similar to UNet \cite{ronneberger2015u} is developed to detect articulation joint of the surgical instrument \cite{colleoni2019deep}. Nonetheless, more works are needed for real-time application using high-resolution images.

\subsubsection{Multi-task learning (MTL)}

Multi-task learning models in surgical instrument tracking is still an open challenge. There are very few models that are trying to tackle the multiple tasks within a single network. A concurrent instrument segmentation and pose estimation multi-task learning (MTL) method \cite{laina2017concurrent} is designed with ResNet-50 \cite{he2016deep} as encoder, FCN decoder and a regression block of fully-connected layer. However, the model is not suitable for the real-time application, and parts segmentation performance is not satisfactory. In computer vision, several MTL models such as MaskRCNN \cite{he2017mask}, Blitznet \cite{dvornik2017blitznet}, and Sistu et al. \cite{sistu2019real} are developed for joint semantic segmentation and object detection tasks. More recently, UberNet \cite{kokkinos2017ubernet} and Islam et al. \cite{islam2019learning} are tried to optimize MTL model with the multi-phase training approach and tuning. Nonetheless, lower inference speed and optimizing both tasks in a convergence point are still the open challenges in the MTL model.

\subsection{Contributions}
In this work, we address the problems of real-time MTL model for robotic instrument detection and segmentation with asynchronous task-aware optimization and attention pruning. Our contributions are summarized as follows:
\begin{itemize}
    \item[--] We propose a real-time MTL model with a light weight-shared encoder and task-aware spatial decoders for detection and segmentation.
    \item[--] We introduce a novel way to train proposed MTL model by using asynchronous task-aware optimization (ATO) technique. 
    \item[--] To reduce model computation and singularity, we design an attention-based pruning method known as global attention dynamic pruning for trained models with dynamic learning.
    \item[--] We develop an innovative design of segmentation decoder with fusing bypass connection in scSE to boost up the model performance.
    \item[--] We annotate the bounding box for the instruments types with robotic instrument segmentation challenge \cite{allan20192017} dataset. Our model achieves impressive results in detection and surpasses the existing state-of-the-art segmentation models and participant methods of the challenge.
\end{itemize}

\section{Methods} 
In this work, we construct a real-time multi-task learning model with light-weight encoder of VGG16\cite{simonyan2014very} and Skip-scSE decoder, SSD decoder\cite{liu2016ssd} for segmentation and detection tasks respectively. To optimize our MTL model at the same convergence point, we introduce asynchronous task-aware optimization technique (ATO).  We also designed a novel skip spatial-channel squeeze and excitation module (Skip-scSE), by integrating skip connection to enhance segmentation prediction. Moreover, to boost real-time computation, reduce singularity, and redundancy in the model, we propose global attention dynamic pruning (GADP) method.

\subsection{Skip-scSE} 
\begin{figure}[!h]
\centering
\includegraphics[width=0.5\textwidth]{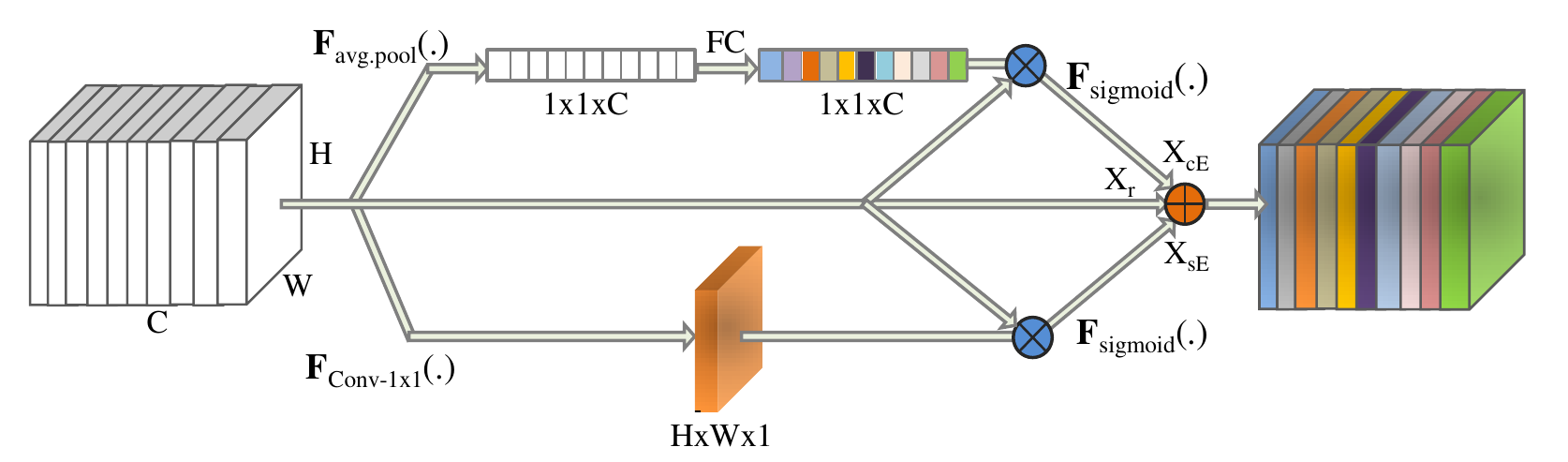}
\caption{Our proposed Skip spatial-channel squeeze and excitation model. Skip eliminates the singularity and sparsity in the CNN learning.}
\label{fig:scse}
\end{figure}

Spatial and channel squeeze \& excitation (scSE) \cite{roy2018concurrent, hu2018squeeze} is a parameter learning technique which learns by recalibrates the feature maps to suppress the weak features and signify the meaningful features. However, in deeper layers, it may cause sparsity and singularity due to squeezing of irrelevant features towards zero \cite{uhrig2017sparsity}. In \cite{orhan2017skip}, skip connection is used to eliminates the singularity and sparsity in the CNN learning. To resolve this issue, we also integrate skip connection which retains the weak features and enhance the useful features by exciting them. We propose skip-scSE unit, by adding a bypass connection with scSE as in Fig. \ref{fig:scse}. In skip-scSE, the output of channel excitation $\tilde{\mathrm{x}}_{cE}$ and spatial excitation $\tilde{\mathrm{x}}_{sE}$ are fused with skip input $\mathrm{x}_{r}$. Therefore, the output excited feature maps $\mathrm{x}_{SC}$ can be formulated as given in equation \ref{equ:skip-scSE},

\begin{eqnarray}
\label{equ:skip-scSE}
\mathrm{x}_{S} &=& \tilde{\mathrm{x}}_{cE} + \tilde{\mathrm{x}}_{sE} + \mathrm{x}_{r} 
\end{eqnarray}

\subsection{Network Architecture}
\begin{figure}[!h]
\centering
\includegraphics[width=0.45\textwidth]{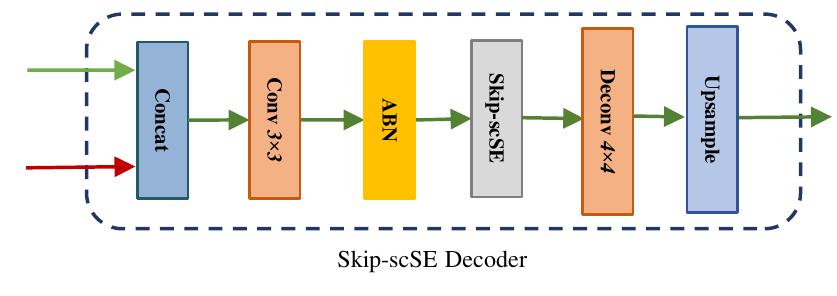}
\caption{Our proposed Skip-scSE decoder for segmentation of surgical iinstruments.}
\label{fig:seg_decoder}
\end{figure}

Our AP-MTL model forms of a shared-encoder, a segmentation decoder, and a detection block (illustrated in Fig. \ref{fig:net_arch}). The encoder and detection block adopt from SSD \cite{liu2016ssd} where VGG16 \cite{simonyan2014very} exploit in the encoder section. The segmentation decoder is designed with proposed skip-scSE by following our previous work \cite{islam2019learning}, as shown in Fig. \ref{fig:seg_decoder}.  Multi-scale feature maps are extracted from different stages in the encoder network. The score maps the feature network are concatenated with high-level score maps to increase parameter learning, and then it gets convoluted and excited. Further excited score maps are upsampled with a deconvolution layer. The final semantic score map will be generated after the last upsampling, which is used to output the prediction results. 

\subsection{Asynchronous Task-aware Optimization (ATO)} 
The fundamental problem in optimizing an MTL model is obtaining optimal minima at the same convergence point. In prior approaches, optimization of multiple tasks uses a naive weighted sum of losses \cite{zhang2017survey, he2017mask, dvornik2017blitznet}, where the loss weights are uniform, or manually tuned. This kind of optimization can be related as optimizing a single task with added noise, where the noise correlates other task losses. To overcome these problems, we introduce Asynchronous Task-aware Optimization (ATO). In this technique, we optimize the task-aware spatial decoder and detection block by calculating gradient independently. Further, we simultaneously regularize end-to-end MTL model to attain a global convergence point for a few epochs with low learning rate.

\begin{algorithm}[!h]
\caption{\small{Asynchronous Task-aware Optimization}}
\begin{algorithmic}[1]
\label{algortihm_ato}
\small
\STATE \textbf{Initialize model weights}:\\shared (${W_{sh}}$), detection (${W_d}$),  segmentation decoder (${W_s}$)\\
\STATE \textbf{Set gradient accumulators to zero:}\\shared (${dW_{sh}}$), detection (${dW_d}$),  segmentation (${dW_s}$)\\
 $\mathbf{dW_{sh}} \leftarrow 0,\; \mathbf{dW_{d}} \leftarrow 0,\; \mathbf{dW_{s}} \leftarrow 0$\\
\STATE \textit{[Optimize Detection]}\\
 $\mathbf{while}\; detection\; task\; not\; converged\; \mathbf{do}:$ \\
\hspace{0.25cm}\textit{[Shared encoder gradients w.r.t detection loss ${\L_{d}}$]}\\
 \hspace{0.25cm}$\mathbf{dW_{sh}} \leftarrow \mathbf{dW_{sh}} + \sum_{i}\delta_{i}\nabla_{W_{sh}}\L_{d}(W_{sh},W_{d})$\\
 \hspace{0.25cm}\textit{[detection block gradients w.r.t detection loss ${\L_{d}}$]}\\
 \hspace{0.25cm}$\mathbf{dW_{d}} \leftarrow \mathbf{dW_{d}} + \sum_{i}\delta_{i}\nabla_{W_{d}}\L_{d}(W_{sh},W_{d})$\\
 $\mathbf{end \; while}$\\
\STATE \textit{[Optimize Segmentation]}\\
 $\mathbf{while}\; segmentation\; task\; not\; converged\; \mathbf{do}:$ \\
 \hspace{0.25cm}\textit{[segmentation gradients w.r.t segmentation loss ${\L_{s}}$]}\\
\hspace{0.25cm}$\mathbf{dW_{s}} \leftarrow \mathbf{dW_{s}} + \sum_{i}\delta_{i}\nabla_{W_{s}}\L_{s}(W_{sh}, W_{s})$\\
 $\mathbf{end \; while}$\\

\STATE \textit{[Regularization]}\\
 $\mathbf{while}\; both\; tasks\; improving\; \mathbf{do}:$ \\
\hspace{0.25cm}$\mathbf{dW_{sh}} \leftarrow \mathbf{dW_{sh}} + \sum_{i}\delta_{i}\nabla_{W_{sh}}\L(W_{sh},W_{d},W_{s})$\\
\hspace{0.25cm}$\mathbf{dW_{d}} \leftarrow \mathbf{dW_{d}} + \sum_{i}\delta_{i}\nabla_{W_{d}}\L(W_{sh},W_{d},W_{s})$\\
\hspace{0.25cm}$\mathbf{dW_{s}} \leftarrow \mathbf{dW_{s}} + \sum_{i}\delta_{i}\nabla_{W_{s}}\L(W_{sh},W_{d},W_{s})$\\
 $\mathbf{end \; while}$\\
\end{algorithmic}
\end{algorithm}

In algorithm \ref{algortihm_ato}, we do Xavier initialization and construct mini-batches for training. At first, the model is trained with detection loss only by detaching segmentation decoder as detecting bounding box is more sensitive task than pixel-wise segmentation. Then, detach the detection block and train only segmentation decoder on segmentation loss by using encoder features trained for the detection task. Encoder features of detection task are also significant for the segmentation as both tasks are targeted on the same objects. Finally, to generalize the shared weights of the encoder, we opt end-to-end regularization by optimizing encoder and both detection block and segmentation decoder with small learning rate \(10^{-5}\). This ensures the MTL model to be more generalized and smooth gradient flow throughout the network.

\subsection{Global Attention Dynamic Pruning (GADP)} 
MTL methods are generally heavy in computation and not applicable in real-time, especially for high-resolution images. Every network contains a sub-network, and most networks' parameter learning is redundant and insignificant. Besides, redundancy and singularity in the learning parameters are the huge hindrances for task-specific decoders in MTL model. Thus, pruning can solve this over-parameterization and singularity problem. In prior works, network are pruned using Taylor expansion ranking\cite{molchanov2016pruning} or attention statistics\cite{yamamoto2018pcas}. Nonetheless, pruning filters based on local gradient flow or activation is highly redundant, as it is a function of all previous filters. Therefore, we propose a dynamic pruning method based on global understanding of parameters and channel-wise attention\cite{hu2018squeeze, wang2017residual}. To avoid initial layer bottleneck pruning, we follow proportional pruning in each layer.

\begin{figure}[!h]
\centering
\includegraphics[width=0.45\textwidth]{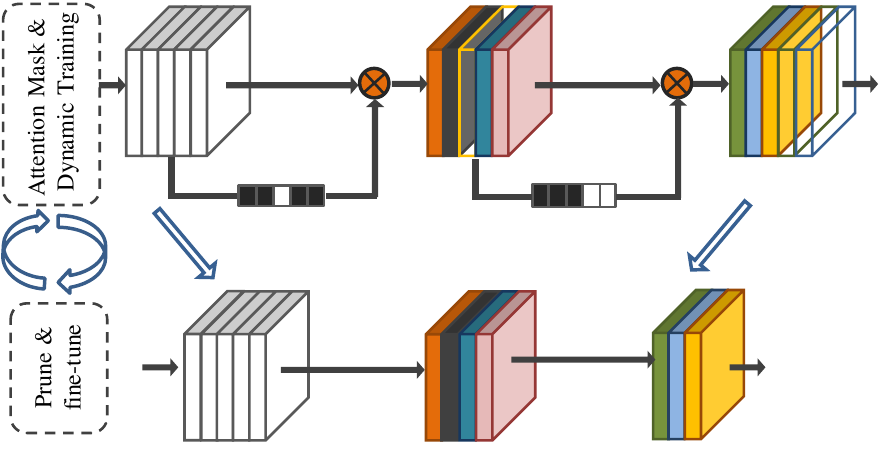}
\caption{Overview of our Global Attention Dynamic Pruning (GADP) method. GADP attaches in-between encoder blocks, calculate the rank of the each kernel, and remove low ranked kernels gradually by retraining the model. Attention-based pruning, rank the network parameters accurately which assists to remove redundant and weak parameters.}
\label{fig:ATO}
\end{figure}

\begin{algorithm}[!h]
\caption{\small{Global Attention Dynamic Pruning}}
\begin{algorithmic}[1]
\label{algortihm_gadp}
\small
\STATE \textbf{Input:} The pre-trained model $\mathcal{M}$, the training set $\mathcal{D}$, the testing set \textit{T}, Accuracy, Drop \(d\),  attention iter \textit{I}, finetune iter \textit{N}.\\
\STATE \textbf{Ouput:} The pruned model $\mathcal{M}'$.\\
 \(T_m\) $=$ Accuracy for pre-trained model $\mathcal{M}$ with testing set \textit{T}\\
 Attach SE-Attention modules to the conv layers of $\mathcal{M}$.
\STATE \textbf{for} attention iteration $i = 1, 2,\ldots, I$ \textbf{do}\\
 \hspace{0.25cm} Forward propagation with $\mathcal{D}$.\\
 \hspace{0.25cm} Update only attention module in backpropagation.\\
 $\mathbf{end \; for}$\\
\STATE Calculate the attention statistics $a_{l,c}$ with $\mathcal{D}$.\\
 Threshold ($t$) according to proportional pruning strategy\\ 
 \textbf{for} {each layer $l=1,2,\ldots, L$} \textbf{do}\\
 \hspace{0.25cm} \textbf{for} {each channel $c=1,2,\ldots, C_l$} \textbf{do}\\
 \hspace{0.5cm} \textbf{if} {$a_{l,c} < t/C_l$} \textbf{then}\\
 \hspace{0.75cm}  mask the channel $c$ from $\mathcal{M}$.\\
 \hspace{0.5cm} \textbf{end if}\\
 \hspace{0.25cm} \textbf{end for}\\
 $\mathbf{end \; for}$
\STATE \textbf{for} finetune iteration $i = 1, 2,\ldots, N$ \textbf{do}\\
 \hspace{0.25cm} Forward propagation with $\mathcal{D}$.\\
 \hspace{0.25cm} Update only masked model $\mathcal{M}'$  in backpropagation.\\
 \textbf{end for}\\
\STATE \(T_m^{'}\) $=$ Accuracy for masked model $\mathcal{M}'$ with testing set \textit{T}\\
 \textbf{if} \(T_m\)$-$\(T_m^{'}\) $<$ \(d\) \textbf{then}\\
 \hspace{0.50cm}  \textbf{goto} Step 3:\\
 \textbf{else}\\
 \hspace{0.25cm} pruned model $\mathcal{M}'$ = prune masked channel in $\mathcal{M}'$\\
 \hspace{0.25cm} \textbf{for} prune regularize iteration $i = 1, 2,\ldots, N$ \textbf{do}\\
 \hspace{0.5cm} Forward propagation with $\mathcal{D}$.\\
 \hspace{0.5cm} Update pruned model $\mathcal{M}'$ parameters in backpropagation.\\
 \hspace{0.25cm} \textbf{end for}\\
 \hspace{0.25cm}  \textbf{return} pruned model $\mathcal{M}'$\\
 \textbf{end if}
\end{algorithmic}
\end{algorithm}

In algorithm \ref{algortihm_gadp}, first, we append SE-Attention module to our pre-trained MTL model's encoder. Then, for a few epochs, we train only attention module weights to learn a better global understanding of network filters correlation. Further, a pruning plan is generated with trained attention statistics by setting the mask threshold conditioned upon proportional pruning strategy. Then, we detach the attention modules and dynamically fine-tune the MTL model by masking channel without pruning. Repeat the whole process until our masked model accuracy drops below the specified margin with the required number of parameters to be masked. Finally, prune all masked channels to obtain an efficient weight shared encoder and regularize it for few epochs for better gradient flow and generalization. Overview of our proposed GADP is illustrated in Fig. \ref{fig:ATO}.

\section{Experiments}

\begin{table*}[!h]
\centering
\caption{Evaluation score for the cross-validation dataset. The evaluation metrics are dice, Hausdorff(Hausd.), mean Average Precision(mAP), frame per second (FPS) and parameters(Param). FPS is calculated by single RTX 2080 Ti GPU with batch size of 1. The best values of each metric are boldened and the values better than ours are underlined.}
\label{table:overall}
\begin{tabular}{|l|c|c|c|c|c|c|c|c|}
\hline
\multicolumn{1}{|c|}{\multirow{1}{.8cm}{\textbf{Model}}} & \multirow{1}{.8cm}{\textbf{Regime}} & \multicolumn{4}{c|}{\textbf{Segmentation}} & \textbf{\begin{tabular}[c]{@{}c@{}} Obj. Detect. \end{tabular}} & \multicolumn{2}{l|}{\textbf{General}} \\ \cline{3-9} 
\multicolumn{1}{|c|}{} &  & \multicolumn{2}{c|}{\textbf{Binary}} & \multicolumn{2}{c|}{\textbf{Type}} & \multirow{1}{.6cm}{\textbf{mAP}} & \multirow{1}{.6cm}{\textbf{FPS}} & \multicolumn{1}{l|}{\multirow{1}{.8cm}{\textbf{Param}}} \\ \cline{3-6}
\multicolumn{1}{|c|}{} &  & \textbf{Dice} & \textbf{Hausd.} & \textbf{Dice} & \textbf{Hausd.} &  &  & \multicolumn{1}{l|}{} \\ \hline
{Ours} & {Seg. \& Detect.} & \textbf{0.947} & \textbf{9.469} & \textbf{0.704} & 10.56 & 0.392 & {18} & \multicolumn{1}{l|}{22.4M} \\ \hline
MaskRCNN \cite{he2017mask} & Seg.\& Detect. & 0.452 & - & 0.283 & - & 0.376 & 4 & - \\ \hline
SSD \cite{liu2016ssd} & Detect. & - & - & - & - & \underline{0.406} & 12 & \multicolumn{1}{l|}{22.6M} \\ \hline
YoloV3 \cite{redmon2018yolov3} & Detect. & - & - & - & - & 0.363 & 12 & - \\ \hline
LinkNet \cite{chaurasia2017linknet} & Seg. & 0.941 & 10.39 & 0.578 & 11.88 & - & 111 & \multicolumn{1}{l|}{11.5M} \\ \hline
ERFNet \cite{romera2017erfnet} & Seg. & 0.923 & 11.05 & 0.495 & 11.55 & - & 44 & \multicolumn{1}{l|}{\underline{2.02M}} \\ \hline
DeepLabv3+ \cite{chen2017rethinking} & Seg. & \underline{0.947} & 10.07 & 0.657 & \underline{10.47} & - & 56 & 59.3M \\ \hline
scSE UNet \cite{roy2018concurrent} & Seg. & 0.931 & 10.09 & 0.542 & 11.88 & - & 107 & 42.7M \\ \hline
TernausNet11 \cite{iglovikov2018ternausnet} & Seg. & 0.920 & 11.25 & 0.446 & 12.29 & - & \underline{295} & 34.5M \\ \hline
BiseNet \cite{yu2018bisenet} & Seg. & 0.942 & 9.997 & 0.440 & 10.80 & - & 54 & 90.8M \\ \hline
GCN \cite{peng2017large} & Seg. & 0.940 & 10.48 & 0.612 & 13.98 & - & 83 & 23.9M \\ \hline
\end{tabular}
\end{table*}

\subsection{Dataset} 

\begin{figure}[!h]
\centering
\includegraphics[width=.45\textwidth]{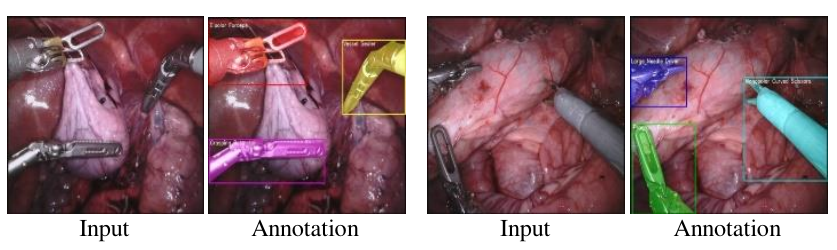}
\caption{Dataset visualization of two random frames. We annotate the bounding box on MICCAI robotic instrument segmentation dataset \cite{allan20192017}.}

\label{fig:data_visual}
\end{figure}

This work is done using robotic instrument segmentation dataset of MICCAI endoscopic vision challenge 2017 \cite{allan20192017}. This dataset contains 10 sequences recorded with the resolution of 1920 x 1080 using Da Vinci surgical systems\cite{ngu2017vinci}. In each sequence, significant instrument motion and visibility is observed and sampled at a rate of 1 Hz. 
The training set consist of sequence 1, 2, 3, 5, 6, 8, and cross-validation data consist of sequence 4, 7, and testing data consist of sequence 9, 10. For the training and cross-validation data sequence, we crop the frames into 1280 x 1024 by removing black padding and annotate bounding box for the instrument with the aid of instrument type segmentation as shown in Fig. \ref{fig:data_visual}.

\subsection{Implementation details}
The input images are normalized by subtracting the image mean and dividing standard deviation. While training, we sequentially implemented algorithm \ref{algortihm_gadp} and algorithm \ref{algortihm_ato}. For other hyper-parameters, we use SGD optimizer with an initial learning rate 0.0001 and `poly' learning rate with the power of 0.9 to update it. The momentum and weight decay set constant to 0.99 and $10^{-4}$ respectively.

\subsection{Post-processing}
\begin{figure}[!h]
\centering
\includegraphics[width=.45\textwidth]{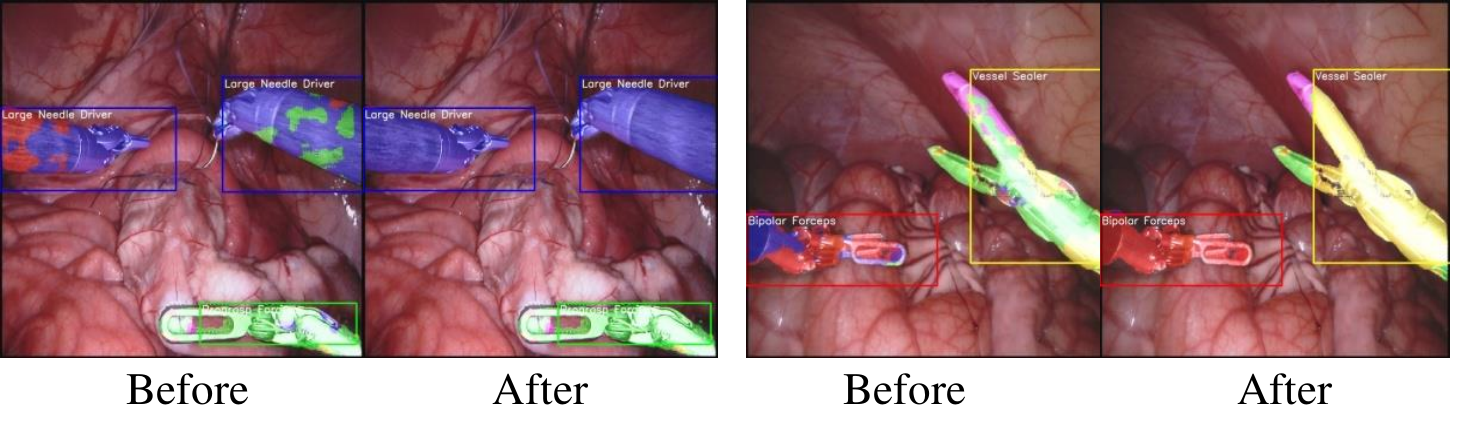}
\caption{Post-processing on predicted segmentation using bounding box. It boosts the type wise segmentation accuracy.}
\label{fig:post_processing}
\end{figure}

In post-processing, we implemented a novel method for our architecture. Quantitatively we observed that our model can perform well in binary segmentation but under-perform in instrument type segmentation. Thus, we inferred that our model can discriminate instruments and tissues but struggles to differentiate between each instrument types. Moreover, we observed that as detection is conditioned upon IOU greater than 0.5, a major area of predicted bounding box covers the instrument\cite{liu2016ssd}. However, in the case of the overlapping bounding box, the area of intersection is not subjected to post-processing. Therefore for high confidence predicted box, we assign all pixels inside the bounding box to its class in predicted instrument type output.  By following this method, instrument type segmentation accuracy improved around 20-25\%. Fig. \ref{fig:post_processing} shows the model prediction before and after post-processing. 

\section{Results and Evaluation}

We evaluate segmentation accuracy of our model using evaluation metrics like dice coefficient, Hausdorff, sensitivity, specificity and for object detection accuracy by mean average precision (mAP) and real-time performance by frames per second (fps). In Table \ref{table:overall}, we compare our architecture with the existing state-of-the-art architectures for segmentation, object detection, and multi-tasking. Inference time for all the models is calculated by using a single NVIDIA RTX 2080 Ti GPU with a batch size of one.

\subsection{Evaluation with Cross-validation}

\begin{figure}[!h]
\centering
\includegraphics[width=.5\textwidth]{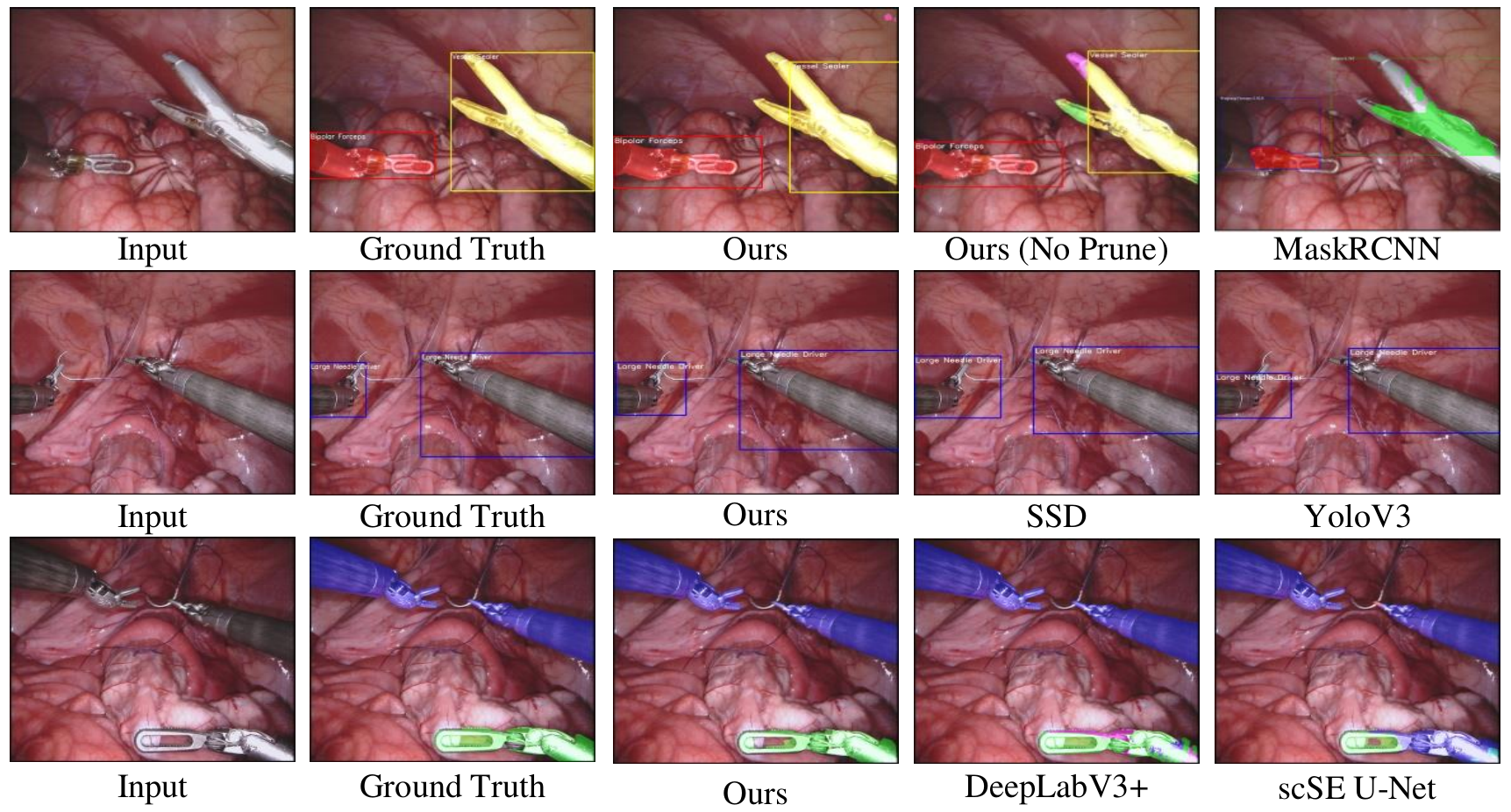}
\caption{The comparison of the type-wise segmentation and detection prediction. Predictions of our model contain lower false positive, false negative and competitive bounding box with SSD \cite{liu2016ssd}.}
\label{fig:type_seg_det}
\end{figure}

From Table \ref{table:overall}, our model outperforms state-of-the-art multi-tasking architecture like MaskRCNN\cite{he2017mask} in both segmentation and object detection. Additionally in instrument type segmentation, our model surpasses single task state-of-the-art architectures such as LinkNet\cite{chaurasia2017linknet}, scSE UNet\cite{roy2018concurrent} and ERFNet\cite{romera2017erfnet} with a large margin. In object detection, our model produces competitive results compared to single-task state-of-the-art network such as SSD\cite{liu2016ssd}and perform better than YoloV3\cite{redmon2018yolov3}. In comparison with MTL model like  MaskRCNN\cite{he2017mask}, our model outperforms type-wise segmentation accuracy. Also, produce a competitive result in object detection with outperforming real-time computation. Further, from the above results, multi-task learning is more generalized and scalable and produce better results than single-task learning\cite{sistu2019real}. Fig. \ref{fig:type_seg_det} demonstrates the type segmentation prediction with our AP-MTL model.

\subsection{Evaluation with Challenge}

\begin{table}[!htbp]
\centering
\caption{Comparison of quantitative results for the binary and type base segmentation tasks of proposed model and reported performance in \cite{allan20192017}. The best values of each metric are boldened. The values better than ours are underlined.}
\label{table:chal}
\begin{tabular}{|l|c|c|c|c|}
\hline
\multicolumn{1}{|c|}{\multirow{1}{1cm}{\textbf{Model}}} & \multicolumn{2}{c|}{\textbf{Binary Seg.}} & \multicolumn{2}{c|}{\textbf{Type Seg.}} \\ \cline{2-5} 
\multicolumn{1}{|c|}{} & \textbf{Dataset9} & \textbf{Dataset10} & \textbf{Dataset9} & \textbf{Dataset10} \\ \hline
\textbf{Ours} & \textbf{0.883} & {0.892} & {0.350} & \textbf{0.795} \\ \hline
MIT & 0.865 & 0.905 & \underline{0.357} & 0.609 \\ \hline
SIAT & 0.839 & 0.899 & 0.315 & 0.791 \\ \hline
TUM & 0.877 & \underline{0.909} & - & - \\ \hline
Delhi & 0.626 & 0.715 & - & - \\ \hline
UCL & 0.808 & 0.869 & 0.272 & 0.583 \\ \hline
NCT & 0.789 & 0.899 & 0.247 & 0.552 \\ \hline
UB & 0.855 & 0.917 & 0.106 & 0.709 \\ \hline
BIT & 0.236 & 0.403 & - & - \\ \hline
UA & 0.539 & 0.689 & 0.040 & 0.715 \\ \hline
UW & 0.377 & 0.603 & - & - \\ \hline
\end{tabular}
\end{table}
 
\begin{figure}[!htbp]
    \centering
    \includegraphics[width=0.48\textwidth]{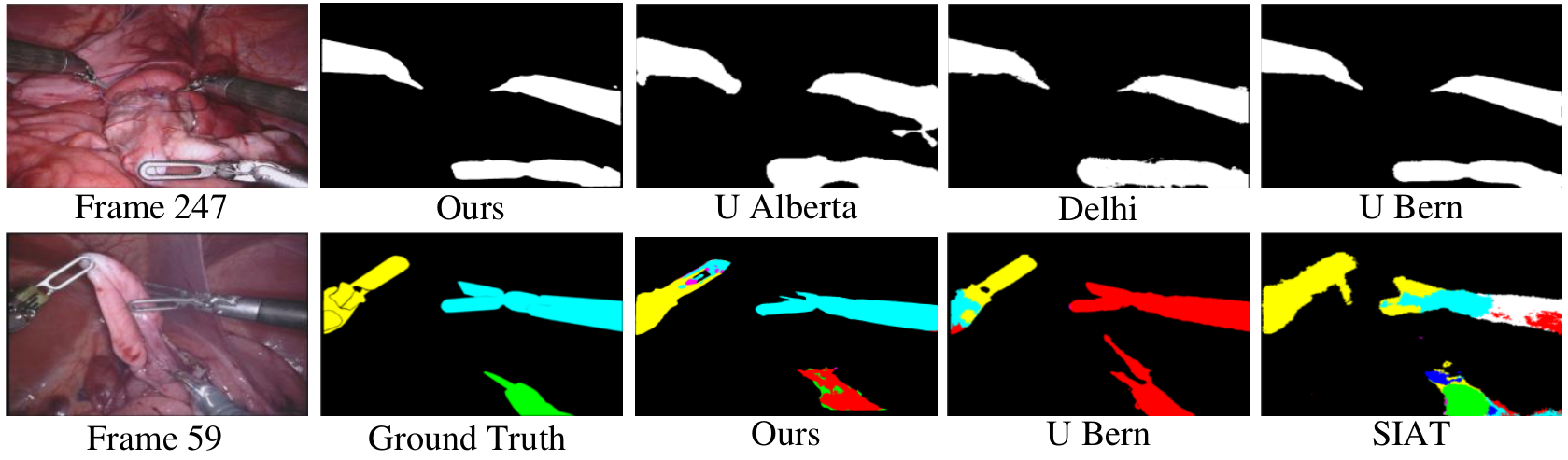}
    \caption{Qualitative results comparison of frames from different dataset for binary and type segmentation with corresponding results from the methods of robotic instrument segmentation challenge \cite{allan20192017} and proposed AP-MTL.}
    \label{fig:chal}
\end{figure}

Table \ref{table:chal} contains a mean intersection-over-union (IoU) performance comparison between our method and challenge methods for binary and type segmentation in MICCAI 2017 robotic instrument segmentation challenge \cite{allan20192017}. Our model prediction of sequences 9 and 10 is evaluated by challenge portal\footnote[1]{https://endovissub2017-roboticinstrumentsegmentation.grand-challenge.org/}. We compare the prediction of our model with the binary and instrument type segmentation of the participated models in the challenge \cite{allan20192017}. From the table, our model achieves competitive performance with all other participants' models for both binary and type segmentation task. From Fig. \ref{fig:chal}, our segmentation predictions of a less false positive and false negative than other challenge models.

\section{Ablation Study}

\subsection{Module and Methods}

To evaluate our method and integration of the modules, we calculate the performance and complexity of the model, as shown in Table \ref{table:module_ab}. Proposed optimization approach, ATO produces a considerable impact on our MTL model for both segmentation and detection tasks. It fixes the common challenge of MTL model to attain optimal minima at a convergence point (at same epoch). We monitor that MTL model without ATO converges the two tasks into different epochs. Without ATO, the convergence epoch of segmentation consists of poor detection accuracy in Table \ref{table:module_ab}.
Moreover, ATO achieves the convergence of both tasks into the same epoch with the best accuracy for both tasks of segmentation and detection. Skip-scSE and GADP enhance the segmentation and detection accuracy with a significant margin, respectively. High-resolution images require high computational resource with larger learning parameters and remarkably reduce the prediction FPS (frame per second). Thus, GADP removes redundant parameters and boost fps. Moreover, MTL regularization smoothes gradient flow throughout the model, which increase generalization and optimized prediction. From Table \ref{tab:ab_scse}, we can infer that our proposed attention model requires same learnable parameters but outperforms scSE\cite{roy2018concurrent}.

\begin{table}[!h]
\centering
\caption{Our model performance and complexity while integrating different proposed methods and modules. Regu. denotes regularization.}
\label{table:module_ab}
\begin{tabular}{|c|c|c|c|c|c|c|}
\hline
\multicolumn{4}{|c|}{\textbf{Modules and Methods}} & \textbf{Seg.} & \textbf{Det.} & \multicolumn{1}{l|}{\multirow{1}{.6cm}{\textbf{FPS}}} \\ \cline{1-6}
\textbf{ATO} & \textbf{Skip-scSE} & \textbf{GADP} & \textbf{Regu.} & \textbf{Dice} & \textbf{mAP} & \multicolumn{1}{l|}{} \\ \hline
\cmark & \cmark & \cmark & \cmark & 0.704 & 0.392 & \textbf{18} \\ \hline
\cmark & \cmark & \cmark & \xmark & 0.685 & 0.392 & \textbf{18} \\ \hline
\cmark & \cmark & \xmark & \xmark & \textbf{0.714} & \textbf{0.406} & 10 \\ \hline
\cmark & \xmark & \xmark & \xmark & 0.695 & \textbf{0.406} & 11 \\ \hline
\xmark & \xmark & \xmark & \xmark & {0.610} & {0.379} & 11 \\ \hline
\end{tabular}
\end{table}

\begin{table}[!h]
\centering
\caption{Experimental comparison between no attention, scSE \cite{roy2018concurrent} and our attention proposed model. From results, we can infer that our attention module enhances segmentation prediction by a significant margin.}
\label{tab:ab_scse}
\begin{tabular}{|l|c|c|c|}
\hline
\multirow{1}{1cm}{\textbf{Modules}} & \multicolumn{1}{l|}{\textbf{Seg.}} & \textbf{Det.} & \multicolumn{1}{l|}{\multirow{1}{.6cm}{\textbf{FPS}}} \\ \cline{2-3}
 & \textbf{Dice} & \textbf{mAP} & \multicolumn{1}{l|}{} \\ \hline
 \textbf{No Attention} & \multicolumn{1}{l|}{0.658} & \multicolumn{1}{l|}{0.406} & 11 \\ \hline
\textbf{scSE} & 0.695 & 0.406 & 11 \\ \hline
\textbf{Skip-scSE} & \textbf{0.714} & 0.406 & 10 \\ \hline
\end{tabular}
\end{table}

\subsection{Single-task and Multi-task}

From the comparison Table \ref{table:single_ab} between single-task and multi-task, the MTL model improves the performance of both tasks comparing to single-task models for segmentation and detection. Our proposed MTL optimization technique, ATO, is playing crucial rule behind the performance improvement. Moreover, MTL regularization refines the learning parameters for both tasks and attains optimal convergence point.

\begin{table}[!h]
\caption{Performance comparison between single-task, multi-task model and regularization. Our multi-task model achieves higher performance than the models of the individual tasks.}
\label{table:single_ab}
\centering
\begin{tabular}{|c|c|c|c|c|c|c|}
\hline
\multicolumn{3}{|c|}{\textbf{Task}} & \multicolumn{2}{c|}{\textbf{Segmentation}} & \textbf{Detection} & \multirow{1}{.6cm}{\textbf{FPS}} \\ \cline{1-6}
\multirow{1}{.6cm}{\textbf{Seg.}} & \multirow{1}{.6cm}{\textbf{Det.}} & \multirow{1}{.6cm}{\textbf{Regu.}} & \textbf{Binary} & \textbf{Type} & \multirow{1}{.6cm}{\textbf{mAP}} &  \\ \cline{4-5}
 &  &  & \textbf{Dice} & \textbf{Dice} &  &  \\ \hline
\cmark & \xmark & \xmark & 0.939 & 0.680 &  & \textbf{56} \\ \hline
\xmark & \cmark & \xmark & - & - & \textbf{0.406} & 12 \\ \hline
\cmark & \cmark & \xmark & 0.941 & 0.685 & 0.392 & 18 \\ \hline
\cmark & \cmark & \cmark & \textbf{0.947} & \textbf{0.704} & 0.392 & 18 \\ \hline
\end{tabular}
\end{table}

\subsection{Prune ration}

\begin{figure}[!h]
    \centering
    \includegraphics[width=0.5\textwidth]{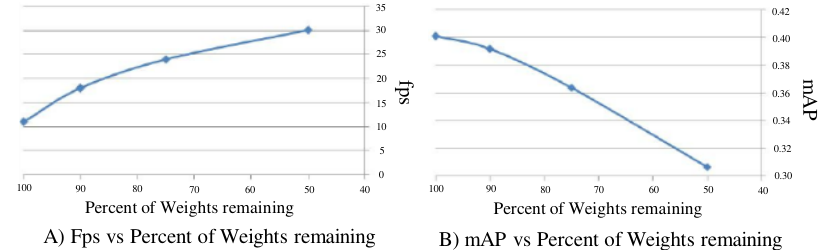}
    \caption{Graphical visualization of A) fps vs Percent of weights remaining B) mAP vs Percent of weights remaining. Total network parameters with 100 \% weights remaining is 22.6 M.}
    \label{fig:comp_graph}
\end{figure}

Variation of model performance with pruning is studied and illustrated in Fig. \ref{fig:comp_graph}. As we increase prune ration, mAP reduces by a small margin with significant improvement in fps. Attention-based pruning aids removal of redundant filters with global receptive field perception. Therefore, noise variants in the model can be reduced by pruning, and detection can be reinforced.

\section{Conclusion}
In this paper, we present attention pruned multi-task learning model (AP-MTL) for real-time instrument localization and tracking in endoscopic surgery. We introduce attention based dynamic pruning technique to eliminate sparsity and singularity in MTL model's shared weight encoder. We present a novel method to optimize the task-aware MTL model to obtain the same optimal convergence point for multi-tasks. To enhance segmentation, we design a novel decoder using skip-scSE to reduce sparsity and redundancy in the decoder. Also, we introduced a novel post-processing method which exploits the benefit of multi-tasking for increasing segmentation accuracy. Further, our model outperforms most of the state-of-the-art architecture by a significant margin. In future work, we extend our research to segment defected tissues along with instruments and learning to exploit the surgical scene representation, compositionality, and reasoning.

\bibliographystyle{IEEEtran}
\bibliography{mybib}

\end{document}